\documentclass[letterpaper, 10 pt, conference]{ieeeconf}  

\IEEEoverridecommandlockouts                              

\usepackage{graphics} 
\usepackage{amsmath} 

\usepackage{booktabs}
\usepackage{multirow}
\usepackage{graphicx}
\usepackage{epstopdf}
\usepackage{pifont}
\usepackage{threeparttable}

\makeatletter
\let\NAT@parse\undefined
\makeatother
\usepackage{xcolor}
\definecolor{rblue}{rgb}{0,0.5,1}
\definecolor{hollywoodcerise}{rgb}{0.96, 0.0, 0.63}
\definecolor{lasallegreen}{rgb}{0.03, 0.47, 0.19}
\definecolor{hanpurple}{rgb}{0.32, 0.09, 0.98}
\definecolor{green(pigment)}{rgb}{0.0, 0.65, 0.31}
\usepackage[pagebackref=true,breaklinks=true,colorlinks,bookmarks=false]{hyperref}
\hypersetup{colorlinks,linkcolor={red},citecolor={hanpurple},urlcolor={magenta}}

\title{\LARGE \bf
Towards Consistent Object Detection via LiDAR-Camera Synergy
}

\author{Kai Luo$^{{2},{\ast}}$, Hao Wu$^{{2},{\ast}}$, Kefu Yi$^{1,}$\textsuperscript{\ding{41}}, Kailun Yang$^{3,4}$, Wei Hao$^{1}$, and Rongdong Hu$^{5}$
\thanks{*Contribute equally, \textsuperscript{\ding{41}}Corresponding author.}
\thanks{$^{1}$School of Traffic and Transportation Engineering, Changsha University of Science and Technology, China {\tt\small corfyi@csust.edu.cn}}%
\thanks{$^{2}$College of Automotive and Mechanical Engineering, Changsha University of Science and Technology, China}%
\thanks{$^{3}$School of Robotics, Hunan University, China}%
\thanks{$^{4}$National Engineering Research Center of Robot Visual Perception and Control Technology, Hunan University, China}%
\thanks{$^{5}$Changsha Intelligent Driving Institute, China}%
\thanks{This work was supported by the National Key Research and Development Program of China under Grant 2022YFC3803700, in part by the Changsha Science and Technology Major Project under Grant kh2202002.}
}

\begin{document}

\maketitle
\thispagestyle{empty}
\pagestyle{empty}

\begin{abstract}
As human-machine interaction continues to evolve, the capacity for environmental perception is becoming increasingly crucial. Integrating the two most common types of sensory data, images, and point clouds, can enhance detection accuracy. Currently, there is no existing model capable of detecting an object's position in both point clouds and images while also determining their corresponding relationship. This information is invaluable for human-machine interactions, offering new possibilities for their enhancement. In light of this, this paper introduces an end-to-end Consistency Object Detection (COD) algorithm framework that requires only a single forward inference to simultaneously obtain an object's position in both point clouds and images and establish their correlation. Furthermore, to assess the accuracy of the object correlation between point clouds and images, this paper proposes a new evaluation metric, Consistency Precision (CP). To verify the effectiveness of the proposed framework, an extensive set of experiments has been conducted on the KITTI and DAIR-V2X datasets. The study also explored how the proposed consistency detection method performs on images when the calibration parameters between images and point clouds are disturbed, compared to existing post-processing methods. The experimental results demonstrate that the proposed method exhibits excellent detection performance and robustness, achieving end-to-end consistency detection. The source code will be made publicly available at \url{https://github.com/xifen523/COD}.
\end{abstract}

\section{Introduction}
Human-machine interaction necessitates the perception of the surrounding environment, and object detection is one of the most commonly employed methods of perception. Depending on the data type, object detection can be categorized into 2D object detection~\cite{lv2023detrs} based on images and 3D object detection~\cite{he2020sassd} based on point clouds, each with broad applications. For certain specialized tasks aimed at achieving higher detection accuracy or robustness, there is also 3D object detection~\cite{focalformer3d} that integrates both images and point clouds.

Although existing detectors~\cite{hu2023ealss,liu2022bevfusion} are powerful, they struggle to establish the correspondence of targets across multiple modalities.
Even when the input data consists of images and point clouds, the final detection result is the object's bounding box in the 3D point cloud, without a corresponding bounding box in the 2D image. To simultaneously obtain the target's location in the 3D point cloud and in the image, further post-processing steps are required. There are typically two implementation methods. The first method utilizes the calibration matrix between the point cloud and the camera to calculate the positions of the eight corners of the 3D bounding box in the image and then infers the largest bounding box based on these eight corners. The second method involves using a 2D image detector to match the 2D detection results of the image with the 3D detection results, thereby obtaining both 2D and 3D detection outcomes for the object.

\begin{figure}[!t]
    \centering
    \includegraphics[scale=0.32]{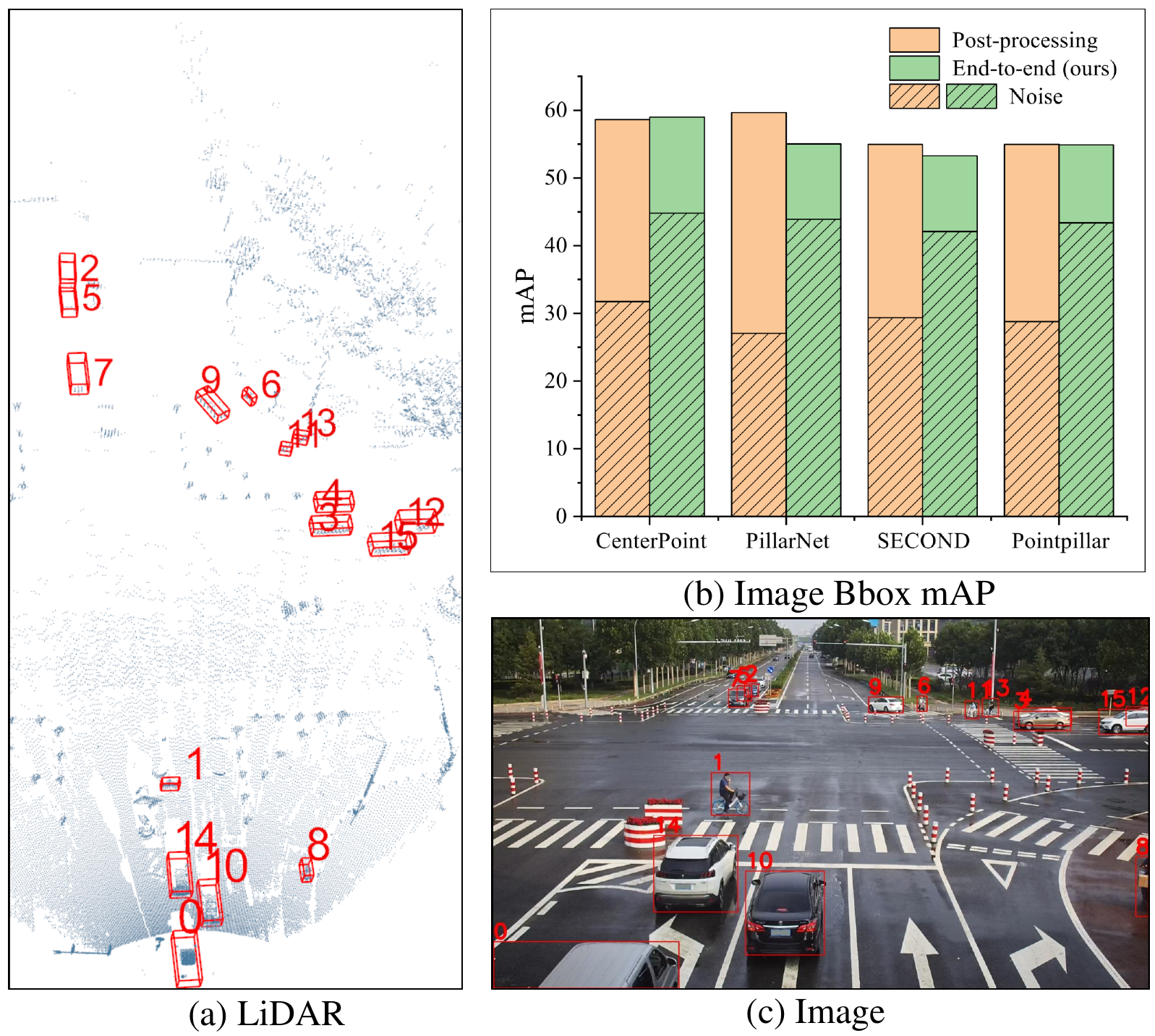}
    \caption{(a) and (c) demonstrate the requirement in consistency detection to simultaneously detect the position of an object in both point clouds and images, with the same object marked with the same ID in both modalities. (b) demonstrates the precision of bounding box detection in images on the KITTI dataset for both the original method and the consistency detection method (ours), under both noisy and noise-free conditions, with the latter showing enhanced robustness.}
    \label{fig:chart}
    \vspace{-0.9cm} 
\end{figure}

To our knowledge, currently, no detector has achieved the capability of obtaining both 2D and 3D detection results in a single inference while ensuring that both results correspond to the same target. 
To address this gap, this paper proposes the task of consistency detection, which involves simultaneously detecting the bounding boxes of an object in different modalities, while ensuring that the detection results represent the same object and are not disordered or confused. 
In addition to existing evaluation methods, this paper introduces the metric of Consistency Precision (\textbf{CP}) to assess the performance of the detector. This metric is used to evaluate whether the objects detected across multiple modalities are indeed the same.

Furthermore, this paper proposes an end-to-end consistency object detection framework based on point clouds and images. It comprises two components: a 3D point cloud detector and a 2D image detector, with the point cloud detector being arbitrary and the image detector specified as a DETR~\cite{Carion2020EndtoEndOD} paradigm detector. In the consistency detection method, the core approach involves using the 3D detection boxes to provide proposals for initializing the queries of the 2D detector, thereby ensuring that the detections by the 2D detector correspond to the proposals from the 3D boxes.
The framework has been verified on the KITTI~\cite{KITTI} and DAIR-V2X~\cite{v2x} datasets, and the results indicate that our method possesses greater robustness compared to existing post-processing methods that calculate 2D bounding boxes using calibration matrices, as shown in Figure~\ref{fig:chart}(b). Additionally, this paper establishes benchmarks for consistency detection on the KITTI~\cite{KITTI} and DAIR-V2X~\cite{v2x} datasets, facilitating research by other researchers in the future.

The main contributions of this paper can be summarized in three aspects:
\begin{itemize}
    \item The task of consistency detection is introduced to exploit LiDAR-camera synergy, along with the metric of consistency precision, for driving scene understanding.
    \item An end-to-end framework for consistency detection is proposed, and benchmarks have been established on the KITTI and DAIR-V2X datasets.
    \item The effectiveness of the proposed framework is verified, showing that it possesses greater robustness across several existing 3D algorithms. This offers a new solution approach for calibration parameter inaccuracies.
\end{itemize}

\section{Related Work}
\subsection{Visual Object Detection}
Visual object detection has evolved significantly. Traditional techniques with manual feature extraction suffered from poor robustness and high computational complexity. Deep learning methods, such as Fast R-CNN~\cite{gr2015frnn}, Faster R-CNN~\cite{Ren2015FasterRT}, and their improved versions~\cite{2017FPN,2017mrcnn}, have achieved excellent results as two-stage detection approaches. Single-stage detectors predict object categories and locations directly from the image, offering speed and lower computational demands.

SSD~\cite{Liu2015SSDSS} was the first single-stage method to achieve real-time performance with accuracy comparable to Faster R-CNN. Following this, YOLO-based methods~\cite{YOLOv3, YOLOv7} have further optimized single-stage detection.

Recent work has introduced end-to-end methods for directly predicting object bounding boxes. DETR~\cite{Carion2020EndtoEndOD} is a key example, using learnable queries to predict object locations and categories. Improved DETR-based methods~\cite{Yao2021EfficientDI,Li2022DNDETRAD,Zhang2022DINODW} continue to emerge, making end-to-end prediction a dominant paradigm by eliminating complex post-processing.

\subsection{Point Cloud Object Detection}
LiDAR sensors can obtain accurate and complete spatial information, but unlike images, they do not yield regular data. 
VoxelNet~\cite{Zhou2017VoxelNetEL} first proposes an end-to-end training network, which is a pioneering work in 3D object detection based on deep learning work. SECOND~\cite{second} proposed sparse convolution methods to reduce memory consumption and increase computational speed. Subsequently, PointPillars~\cite{PointPillar} proposed the idea of encoding point clouds into vertical columns based on VoxelNet to achieve 3D object detection using a 2D object detection framework. 
HVNet~\cite{Ye2020HVNetHV} uses multi-scale voxelization for point cloud processing, by aggregating information from each point within a voxel to compute voxel features and achieve better detection performance on the KITTI~\cite{KITTI} test benchmark.

In addition, the point-based processing is closer to the original LiDAR data and does not lose the original geometric information due to quantization errors. PointNet++~\cite{Qi2017PointNetDH} applies PointNet in a hierarchical recursive manner to improve the detection performance by adaptively capturing the structure and fine-grained features of the point cloud. fine-grained features to improve detection performance. Point RCNN~\cite{Lang2018PointPillarsFE} proposes a point-based two-stage detection framework that extends the classical 2D object detection framework, Faster R-CNN, to 3D object detection. 3DSSD~\cite{Yang20203DSSDP3} proposes a lightweight framework that achieves a balance between detection accuracy and speed.

\begin{figure*}[t!]
    \centering
    \includegraphics[scale=0.58]{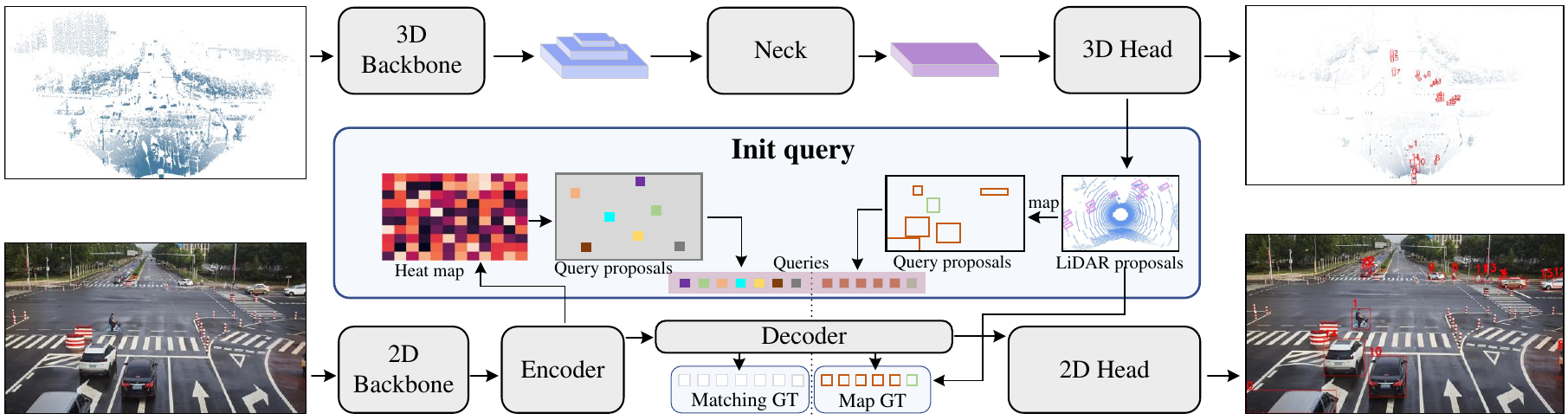}
    \caption{The architecture diagram for the consistency detection network. The overall architecture of consistency detection comprises two pathways: the point cloud object detection pathway and the image object detection pathway. In the former, point cloud features are extracted through a 3D backbone network, transformed via a neck network, and then object positions and dimensions in the point cloud are predicted using a detection head. In the latter, features are extracted through a 2D backbone network and processed through the encoder layer of a transformer to generate a heat map, from which query proposals are derived. Additional query proposals are obtained using the object positions and dimensions acquired from the point cloud. Both sets of query proposals are fed into the decoder layer of the transformer, and the final object positions in the image are obtained via the image's detection head. Notably, during training, the first set of queries is matched with the ground truth to compute loss, while the second set, already corresponding to the ground truth, bypasses the matching process and goes directly to loss calculation.}
    \label{fig:pipeline}
\end{figure*}

\subsection{Fusion-based Object Detection}
The fusion of point clouds and images can compensate for their respective deficiencies and improve perceptual accuracy.    They can be categorized into three types according to the fusion period: early fusion, intermediate fusion, and late fusion.

Early-fusion methods fuse data from different modalities during data preprocessing and typically rely on hard correlations brought about by the transform matrix between the LiDAR and the camera for semantic alignment. PointPainting~\cite{Vora2019PointPaintingSF} is representative of this type of approach, which projects the original point cloud as input to the output of a pure image semantic segmentation network and attaches a category score to each point. MVP~\cite{pan2021variational} utilizes multiple 2D detection networks to generate dense 3D virtual point clouds to augment an otherwise sparse point cloud. However, the sparsity of the points can severely affect the quality of the fusion, so early fusion is not commonly used and a small amount of research work has been done in the early stages of the development of multimodal methods.

Mid-term fusion, which occurs after data preprocessing has been completed and before final detection results are generated, is currently the method with the most potential for development. TransFusion~\cite{Bai2022TransFusionRL} defines object queries in 3D space and fuses image features into these proposal boxes. 
DeepFusion~\cite{Li2022DeepFusionLD} proposes InverseAug inverse rotation and other geometrically relevant data enhancements to achieve precise geometric alignment between LiDAR points and image pixels. SFD~\cite{Wu2022SparseFD} enhances LiDAR-generated sparse point clouds using a dense pseudo-point cloud generated by depth complementation, which fuses 3D RoI features from different point clouds into a 3D mesh. 4D-Net~\cite{Wu2022SparseFD} places the fusion module in the point cloud feature extractor to allow the point cloud features to be dynamically focused on image features.

Post-fusion is also known as decision-level fusion, which focuses on fusing the predictions of different modalities during the decision-making period.
CLOCs~\cite{Pang2020CLOCsCO} is a typical post-fusion method that utilizes maximum suppression (NMS) to post-process the predictions of all modalities. This fusion strategy is based on manual handwritten rules and the post-processing is associated with cumbersome. 

In summary, no existing method can simultaneously detect an object's position in both point clouds and images and establish their corresponding relationship.

\section{Methodology}
\subsection{Network Architecture}
The consistency detection framework proposed in this article integrates a point cloud detection network architecture with the RT-DETR~\cite{lv2023detrs} image detection network, as depicted in Figure~\ref{fig:pipeline}. 
It imposes no constraints on the point cloud detection network, which may either be a single-stage or a two-stage detection network.
The image detection network exclusively employs RT-DETR, chosen for its efficiency and speed, as well as its support for end-to-end image detection. Notably, the architecture achieves target detection in both point clouds and images with a single forward inference, assigning the same ID to the same target. Such end-to-end consistency detection is scarcely achievable using traditional detection methods with NMS post-processing. The essence of consistency detection lies in utilizing the outcomes of point cloud detection as proposals for queries in image object detection, thereby aligning the targets identified in the point clouds with those in the images, as depicted in Figure~\ref{fig:chart}.

\subsection{Learnable Query Initialization}

In the framework proposed in this paper, a learnable query is utilized, which comprises two components: the bounding box (Bbox) position and the category embedding. Initially, a distance calculation is performed between all detected objects from the point cloud and all ground truths to construct a cost matrix. Subsequently, based on this cost matrix, the Hungarian algorithm is employed to associate detected objects with their corresponding ground truths one-to-one. Next, point cloud detections not associated with a ground truth are discarded. The target positions and size from the point cloud detections are then used, along with the calibration matrix between the point cloud and the image, to compute the initial positions of targets in the image. Similarly, the category detected by the point cloud serves as the initial input for the embedding. Notably, the initialized queries here correspond to actual ground truths. These queries can be understood as noisy derivatives of the ground truths, and the subsequent decoder process involves denoising these queries. The initialization of queries based on LiDAR proposals can be represented by the following five equations:
$$cost\_mat = \mathbf{D}(Bbox_\text{LiDAR},GT).  \eqno{(1)} $$    
Here, $cost\_mat$ represents the distance cost matrix. $\mathbf{D}$ denotes the distance calculation function. $Bbox_\text{LiDAR}$ refers to the bounding boxes predicted based on LiDAR data. $GT$ stands for the ground truth values. 
This matrix is used to assess the disparity between each predicted bounding box from LiDAR data and the actual ground truth, facilitating the optimal matching process.
$$ index = \mathbf{Hungarian}(cost\_mat). \eqno{(2)} $$
Here, $index$ represents the indices between the bounding boxes predicted by the LiDAR and the ground truths. $\mathbf{Hungarian}$ denotes the Hungarian matching algorithm. This equation indicates that the Hungarian algorithm is applied to the distance cost matrix $cost\_mat$ to find the best match indices, ensuring that each predicted Bbox from LiDAR is optimally paired with a ground truth entity.
$$Bbox_\text{Image} = \mathbf{Map}(Bbox_\text{LiDAR}^{index}). \eqno{(3)}$$ 
Here, $Bbox_\text{Image}$ denotes the bounding boxes as projected onto the image space. $\mathbf{Map}$ is the mapping function used to translate bounding boxes from the LiDAR coordinate system to the image coordinate system. 
$Bbox_\text{LiDAR}^{index}$ refers to the bounding boxes predicted by the LiDAR that have been successfully matched with ground truth values. This equation illustrates how the bounding boxes, after being matched to ground truths using indices ($index$), are converted from the LiDAR coordinates to image coordinates, effectively mapping them onto the corresponding positions in the image space.
$$ content  = \mathbf{Embedding}(cls^{index}). \eqno{(4)}$$
Here, $\mathbf{Embedding}$ is a function used to transform categorical data into a vector representation. $cls^{index}$ refers to the category associated with each bounding box that has been matched to the ground truths.
The $content$ represents the learnable information derived from the category embeddings. This equation describes how the category associated with each indexed bounding box is transformed into a learnable vector (content) through the embedding function, providing a richer, more informative representation suitable for further processing or learning tasks.
$$qurey=[content, Bbox_\text{image}]. \eqno{(5)}$$
Here, the $query$ thus formed is precisely the initialized query required for subsequent processing. Following the steps outlined above, we can successfully generate queries initialized based on LiDAR proposals.

Additionally, we retain $300$ queries generated from the heat map, consistent with RT-DETR, due to the potential for point cloud omissions that could result in the corresponding targets not being detected in the image. Another advantage of this approach is that the model can still detect targets in the image even without initialization from the point cloud, ensuring it remains functional and robust even in the absence of point cloud inputs. Finally, initializing more queries facilitates rapid model convergence and enhances training efficiency. This is akin to training a parameter-shared network with the same architecture, introducing more supervision, thereby improving DETR training.

\begin{table*}[!t]
\caption{Comparison between the consistency detection method and traditional methods on the KITTI dataset.}
\centering
    \begin{tabular}{lccccccccccccc}
    \toprule
    \multirow{2}{*}{Detector Modality} & \multirow{2}{*}{Modal} & \multicolumn{3}{c}{Car (AP@0.7)} & \multicolumn{3}{c}{Pedestrian (AP@0.5)} & \multicolumn{3}{c}{Cyclist (AP@0.5)} &\multicolumn{3}{c}{mAP}  \\
    \cmidrule(lr){3-5} 
    \cmidrule(lr){6-8}
    \cmidrule(lr){9-11}
    \cmidrule(lr){12-14}
                             &                        & easy  & mod. & hard & easy  & mod. & hard & easy  & mod. & hard &   easy  & mod. & hard          \\
    \midrule
    CenterPoint~\cite{centerpoint}           & \multirow{8}{*}{3D}   & 72.06  & 64.50 & 64.60 & 46.40  & 41.51 & 35.62 & 54.46  & 38.61  & 32.61 & 57.64  & 48.20  & 44.28 \\
    COD (CenterPoint+RT-DETR)   &                       & 72.15  & 69.76 & 64.64 & 46.84  & 42.22 & 41.36 & 52.94  & 36.37  & 30.52 & 57.31  & 49.45  & 45.51 \\
    PillarNet~\cite{pillarnet}             &                       & 58.46  & 58.21 & 53.78 & 38.66  & 32.85 & 31.29 & 56.27  & 37.94  & 33.06 & 51.13  & 43.00  & 39.38 \\
    COD (PillarNet+RT-DETR)     &                       & 58.29  & 59.12 & 54.32 & 38.30  & 32.80 & 31.60 & 56.16  & 38.90  & 37.54 & 50.92  & 43.60  & 41.16 \\
    SECOND~\cite{second}                &                       & 72.97  & 64.70 & 64.18 & 44.85  & 39.10 & 34.36 & 54.39  & 37.21  & 32.21 & 57.40  & 47.00  & 43.58 \\
    COD (SECOND+RT-DETR)        &                       & 75.28  & 65.29 & 64.44 & 44.48  & 38.64 & 33.30 & 51.48  & 35.70  & 34.99 & 57.08  & 46.55  & 44.24 \\
    PointPillar~\cite{PointPillar}           &                       & 65.08  & 61.01 & 55.72 & 35.25  & 30.78 & 25.66 & 52.73  & 37.43  & 36.19 & 51.02  & 43.07  & 39.19 \\
    COD (PointPillar+RT-DETRs)   &                       & 64.91  & 60.94 & 55.52 & 36.87  & 31.77 & 27.53 & 50.78  & 34.84  & 31.27 & 50.85  & 42.52  & 38.11 \\
    \midrule
    CenterPoint~\cite{centerpoint}           & \multirow{8}{*}{Bbox} & 87.33  & 78.50 & 78.42 & 62.51  & 55.43 & 48.44 & 58.99  & 42.02  & 41.81 & 69.61  & 58.65  & 56.22 \\
    COD (CenterPoint+RT-DETR)   &                       & 79.39  & 80.78 & 74.78 & 61.79  & 54.96 & 53.44 & 58.89  & 41.24  & 40.95 & 66.69  & 58.99  & 56.39 \\
    PillarNet~\cite{pillarnet}             &                       & 84.76  & 76.48 & 76.14 & 51.97  & 44.58 & 43.68 & 57.50  & 39.60  & 39.00 & 64.75  & 53.55  & 52.94 \\
    COD (PillarNet+RT-DETR)     &                       & 79.64  & 80.97 & 75.26 & 59.67  & 51.88 & 45.46 & 48.46  & 32.31  & 31.97 & 62.59  & 55.05  & 50.90 \\
    SECOND~\cite{second}                &                       & 88.63  & 86.14 & 79.45 & 58.49  & 52.13 & 45.90 & 64.11  & 40.76  & 39.88 & 70.41  & 59.67  & 55.08 \\
    COD (SECOND+RT-DETR)        &                       & 79.14  & 79.60 & 74.05 & 50.40  & 43.46 & 41.73 & 52.75  & 36.78  & 35.94 & 60.77  & 53.28  & 50.57 \\
    PointPillar~\cite{PointPillar}           &                       & 88.02  & 78.66 & 78.06 & 53.12  & 46.00 & 44.49 & 56.74  & 40.24  & 39.43 & 65.96  & 54.97  & 53.99 \\
    COD (PointPillar+RT-DETR)   &                       & 84.80  & 83.40 & 76.80 & 55.05  & 46.55 & 44.85 & 50.88  & 34.70  & 34.27 & 63.57  & 54.89  & 51.97\\         
    \bottomrule
    \end{tabular}
    \label{tab:consistency_kitti}
\end{table*}

\begin{table*}[!htbp]
\caption{Comparison between the consistency detection method and traditional methods on the DAIR-V2X dataset.}
\centering
    \begin{tabular}{lccccccccccccc}
    \toprule
    \multirow{2}{*}{Detector Modality} & \multirow{2}{*}{Modal} & \multicolumn{3}{c}{Car (AP@0.7)}    & \multicolumn{3}{c}{Pedestrian (AP@0.5)} & \multicolumn{3}{c}{Cyclist (AP@0.5)} & \multicolumn{3}{c}{mAP} \\
    \cmidrule(lr){3-5} 
    \cmidrule(lr){6-8}
    \cmidrule(lr){9-11}
    \cmidrule(lr){12-14}
                             &                        & easy  & mod. & hard & easy  & mod. & hard & easy  & mod. & hard &  easy  & mod. & hard                 \\                      
    \midrule
   CenterPoint~\cite{centerpoint}           & \multirow{8}{*}{3D}   & 71.18  & 62.25 & 62.24 & 13.64  & 13.52 & 13.52 & 42.43  & 41.57  & 41.34 & 42.42  & 39.11  & 39.03 \\
    COD (CenterPoint+RT-DETR)   &                       & 71.39  & 62.48 & 62.46 & 20.81  & 15.06 & 15.15 & 42.93  & 41.59  & 41.46 & 45.04  & 39.71  & 39.69 \\
    PillarNet~\cite{pillarnet}             &                       & 69.85  & 61.28 & 61.24 & 14.46  & 14.22 & 14.22 & 39.40  & 33.00  & 32.90 & 41.24  & 36.16  & 36.12 \\
    COD (PillarNet+RT-DETR)     &                       & 69.75  & 61.36 & 61.32 & 10.03  & 8.89  & 8.89  & 38.64  & 32.68  & 32.53 & 39.48  & 34.31  & 34.25 \\
    SECOND~\cite{second}                &                       & 70.39  & 61.87 & 61.87 & 15.00  & 15.05 & 15.05 & 42.64  & 41.87  & 41.76 & 42.67  & 39.60  & 39.56 \\
    COD (SECOND+RT-DETR)        &                       & 71.02  & 62.26 & 62.24 & 15.70  & 16.00 & 16.00 & 43.14  & 42.09  & 42.05 & 43.28  & 40.12  & 40.10 \\
    PointPillar~\cite{PointPillar}           &                       & 70.52  & 61.66 & 61.59 & 12.05  & 12.47 & 12.47 & 38.48  & 33.18  & 33.10 & 40.35  & 35.77  & 35.72 \\
    COD (PointPillar+RT-DETR)   &                       & 70.16  & 61.50 & 61.42 & 9.38   & 9.66  & 9.66  & 40.02  & 34.22  & 34.16 & 39.85  & 35.12  & 35.08 \\
    \midrule
    CenterPoint~\cite{centerpoint}           & \multirow{8}{*}{Bbox} & 20.09  & 17.51 & 17.50 & 24.56  & 22.74 & 22.70 & 39.84  & 38.89  & 38.91 & 28.16  & 26.38  & 26.37 \\
    COD (CenterPoint+RT-DETR)   &                       & 63.61  & 56.12 & 56.07 & 35.01  & 27.27 & 27.27 & 51.13  & 44.27  & 44.22 & 49.92  & 42.56  & 42.52 \\
    PillarNet~\cite{pillarnet}             &                       & 19.64  & 17.11 & 17.13 & 26.60  & 17.87 & 17.87 & 39.24  & 32.80  & 32.82 & 28.49  & 22.59  & 22.61 \\
    COD (PillarNet+RT-DETR)   &                       & 60.13  & 54.51 & 54.50 & 27.27  & 25.79 & 25.79 & 44.22  & 43.66  & 43.63 & 43.87  & 41.32  & 41.31 \\
    SECOND~\cite{second}                &                       & 19.27  & 17.01 & 17.06 & 25.90  & 24.96 & 24.96 & 39.44  & 39.68  & 39.58 & 28.20  & 27.22  & 27.20 \\
    COD (SECOND+RT-DETR)        &                       & 60.52  & 54.29 & 54.30 & 35.48  & 27.27 & 27.27 & 52.69  & 44.70  & 44.69 & 49.56  & 42.09  & 42.09 \\
    PointPillar~\cite{PointPillar}           &                       & 20.19  & 17.55 & 17.56 & 25.80  & 17.73 & 17.73 & 30.73  & 32.19  & 32.10 & 25.57  & 22.49  & 22.46 \\
    COD (PointPillar+RT-DETR)   &                       & 69.47  & 61.12 & 61.11 & 27.27  & 27.27 & 27.27 & 44.57  & 43.82  & 43.74 & 47.10  & 44.07  & 44.04 \\         
    \bottomrule
    \end{tabular}
    \label{tab:consistency_v2x}
\end{table*}

\subsection{Positive Samples Matching}
In this article, although two sets of queries are concurrently processed through the decoder's attention mechanism, their outputs from the decoder are distinct. Queries initialized based on the heatmap require matching with the ground truth to determine whether they are positive or negative samples. In contrast, queries initialized from point cloud proposals do not require further matching since they are associated with corresponding ground truth from the outset and align with the predicted bounding boxes from the point cloud. This ensures that the targets decoded from the queries correspond to the same targets detected in the point cloud.

\subsection{Training Losses}

The loss function comprises two parts: one from the point cloud detection network and another from the image detection network. To preserve the network's simplicity, we eschew any bells and whistles, simply adding the loss from the point cloud detection to the RT-DETR loss to derive the final loss for optimization.
$$
\mathcal{L} = \mathcal{L}_{\text {LiDAR}}+ \alpha  \mathcal{L}_{\text {Image}}\eqno{(6)}
$$
Here, the term $\mathcal{L}_{\text{LiDAR}}$ represents the loss for the point cloud network, primarily comprising classification and localization losses. Similarly, $\mathcal{L}_{\text{Image}}$ denotes the loss for the image network, consistent with that of RT-DETR. The parameter $\alpha$ serves as a balancing factor between these two losses, and in this study, $\alpha$ is set to $1$.

\section{Experiments}

\subsection{Experimental Setup}
\subsubsection{Datasets}
We primarily evaluated and analyzed our proposed method on KITTI~\cite{KITTI} and DAIR-V2X~\cite{v2x}. The KITTI dataset, a classic object detection dataset, is collected from real traffic scenarios and is particularly valuable because it annotates the spatial positions of objects in both point clouds and image pixels. This dual annotation is advantageous for the training and evaluation of our proposed algorithm. 
Additionally, the DAIR-V2X dataset, also derived from real scenarios, is used in this study. Specifically, we utilize the infrastructure data from this dataset, which similarly annotates the positions of objects in both point clouds and images. It is noteworthy that the original methods compared in this article are capable only of inferring the position of objects within 3D bounding box point clouds. The bounding boxes in images are calculated based on a calibration matrix between the point cloud and the image, utilizing the 3D bounding boxes.

\subsubsection{Metrics}
 We evaluate the detection performance by the mean Average Precision under $11$ recall thresholds (mAP@R11) the same as the official benchmark and evaluation for $3$ classes including Pedestrian, Cyclist, and Car. In the evaluation criteria, if the overlap ratio (Intersection over Union, IoU) of a car's bounding box with the ground truth exceeds $0.7$, it is considered a True Positive (TP). For pedestrians and cyclists, an overlap ratio of $0.5$ is deemed sufficient to classify a detection as TP. This standard aligns with the official evaluation protocols of the KITTI dataset. 

Additionally, this paper introduces a metric for consistency detection called Consistency Precision (CP), which is calculated using the formula: 
$$ CP = \frac{TCD}{GT} $$
Here, $TCD$, or True Consistent Detection, represents the number of 3D detections that correspond to the same target as their 2D detection results, and $GT$ is the number of ground truth samples in 3D detection. The $CP$ value ranges between $0$ and $1$, with higher values indicating better performance of the consistency detection.

\begin{table*}[!t]
\caption{Comparison of consistency precision.}
\centering
\begin{threeparttable}
\resizebox{\linewidth}{!}{
  \renewcommand{\arraystretch}{1.1} %
  
  \begin{tabular}{lcccccccccccc}
    \toprule
    \multirow{3}{*}[-1.4ex]{Detector Modality} & \multicolumn{6}{c}{KITTI} & \multicolumn{6}{c}{DAIR-V2X}   \\
    \cmidrule(lr){2-7}
    \cmidrule(lr){8-13}
    &\multicolumn{2}{c}{Vehicle CP} & \multicolumn{2}{c}{Pedestrian CP} & \multicolumn{2}{c}{Bicycle CP}& 
    \multicolumn{2}{c}{Vehicle CP} & \multicolumn{2}{c}{Pedestrian CP} & \multicolumn{2}{c}{Bicycle CP} \\
    \cmidrule(lr){2-3}
    \cmidrule(lr){4-5}
    \cmidrule(lr){6-7}
    \cmidrule(lr){8-9}
    \cmidrule(lr){10-11}
    \cmidrule(lr){12-13}
                            & noise-free       & noise       & noise-free       & noise        & noise-free       & noise & noise-free       & noise       & noise-free       & noise        & noise-free       & noise       \\
    \midrule
    CenterPoint~\cite{centerpoint} \& RT-DETR~\cite{lv2023detrs}    & 66.45       & 54.3             & 22.11         & 20.2              & 17.36       & 14.33            & 21.34          & 17.55          & 11.85          & 7.47           & 4.73           & 3.69 \\
    COD (CenterPoint+RT-DETR)     & 75.06 & 68.76 & 49.87  & 43.51  & 42.22 & 34.94  & 55.67 & 49.02 & 25.25 & 15.97 & 43.46 & 25.75\\            
    PillarNet~\cite{pillarnet} \& RT-DETR~\cite{lv2023detrs}      & 64.87       & 54.2             & 17.72         & 16.62             & 14.56       & 12.71             & 20.64          & 16.8           & 9.79           & 7.98           & 6.15           & 4.97\\
    COD (PillarNet+RT-DETR)      & 79.17 & 73.88 & 42.72  & 38.29  & 33.59 & 29.68 & 55.46 & 48.28 & 20.1  & {15.97} & {39.77} & {24.24}  \\
    SECOND~\cite{second} \& RT-DETR~\cite{lv2023detrs}         & 68.63       & 57.00               & 20.13         & 18.60              & 15.90        & 13.55           & 19.92          & 16.37          & 10.3           & 9.02           & 6.25           & 5.63 \\ 
    COD (SECOND+RT-DETR)         & {76.21} & {69.98} & {40.00}     & {34.61}  & {38.75} & {33.59}  & {56.26} & {48.28} & {23.71} & {17.01} & {43.65} & {29.82}\\
    PointPillar~\cite{PointPillar} \& RT-DETR~\cite{lv2023detrs}    & 66.08       & 54.89            & 19.82         & 18.44             & 16.01       & 13.89           & 20.52          & 16.66          & 10.82          & 8.76           & 6.53           & 5.2            \\
    COD (PointPillar+RT-DETR)     & {80.22} & {71.88} & {39.82}  & {36.62}  & {37.51} & {32.7}   & {64.25} & {57.39} & {19.58} & {15.46} & {39.58} & {24.24}\\
    \bottomrule
    \end{tabular}
    }
    
\label{tab:comparison_traditional_improved}

    \begin{tablenotes}    %
    \scriptsize               %
        \item[1] ``3D Detector \& RT-DETR'' is a post-processing method where 3D detects 3D boxes and RT-DETR detects 2D boxes. Through matching post-processing techniques, the 
        
        correspondence between the 3D and 2D boxes is established.           %
        \item[2] ``COD (3D Detector+RT-DETR)'' is an end-to-end consistency detection method that we have proposed.         %
    \end{tablenotes}            %
    \end{threeparttable}       %
\end{table*}

\begin{table*}[!t]
\caption{Noise resistance ablation experiments on the KITTI dataset.}
\centering
\resizebox{\linewidth}{!}{
    \begin{tabular}{lccccccccccccc}
    \toprule
     \multirow{2}{*}{Detector Modality} & \multirow{2}{*}{Training Noise} & \multicolumn{3}{c}{Car (Bbox AP@0.7)}    & \multicolumn{3}{c}{Pedestrian (Bbox AP@0.5)} & \multicolumn{3}{c}{Cyclist (Bbox AP@0.5)} & \multicolumn{3}{c}{Bbox mAP} \\
    \cmidrule(lr){3-5}  
    \cmidrule(lr){6-8}
    \cmidrule(lr){9-11}
    \cmidrule(lr){12-14}
                                &                        & easy  & mod. & hard & easy  & mod. & hard & easy  & mod. & hard & easy  & mod. & hard                    \\
    \midrule
        CenterPoint~\cite{centerpoint}             & -             &37.14 & 29.55 & 30.52 & 44.59 & 39.42 & 34.62 & 38.69 & 26.35 & 26.57 & 40.14 & 31.77 & 30.57 \\
        COD (CenterPoint+RT-DETR)   & \ding{55}                         &59.80  & 54.47 & 50.48 & 52.24 & 50.01 & 43.88 & 52.39 & 29.98 & 30.21 & 54.81 & 44.82 & 41.52 \\
        COD (CenterPoint+RT-DETR)   & \ding{51}                         &70.46 & 68.60  & 62.95 & 49.39 & 43.03 & 41.45 & 51.98 & 35.91 & 30.12 & 57.28 & 49.18 & 44.84 \\
       \cmidrule{1-14}
        PillarNet~\cite{pillarnet}                 & -             &28.19 & 22.7  & 24.8  & 38.53 & 32.64 & 31.98 & 40.49 & 25.81 & 25.79 & 35.74 & 27.05 & 27.52 \\
        COD (PillarNet+RT-DETR)     & \ding{55}                         &67.24 & 58.73 & 58.84 & 51.82 & 43.98 & 42.25 & 44.26 & 28.94 & 28.32 & 54.44 & 43.88 & 43.14 \\
        COD (PillarNet+RT-DETR)     & \ding{51}                         &79.74 & 80.95 & 75.26 & 59.93 & 52.07 & 45.63 & 48.55 & 32.31 & 31.97 & 62.74 & 55.11 & 50.95 \\
        \cmidrule{1-14}
        SECOND~\cite{second}                   & -                 &34.62 & 26.59 & 27.63 & 38.35 & 33.55 & 32.69 & 42.07 & 27.97 & 27.7  & 38.35 & 29.37 & 29.34 \\
        COD (SECOND+RT-DETR)            & \ding{55}                     &63.35 & 57.63 & 57.25 & 45.94 & 39.34 & 34.01 & 48.52 & 29.28 & 28.97 & 52.60 & 42.08 & 40.08 \\
        COD (SECOND+RT-DETR)            & \ding{51}                     &75.69 & 77.73 & 72.87 & 52.35 & 44.92 & 43.21 & 42.64 & 26.8  & 26.18 & 56.89 & 49.82 & 47.42 \\
        \cmidrule{1-14}
        PointPillar~\cite{PointPillar}            & -              &35.89 & 28.98 & 29.68 & 36.74 & 31.44 & 30.29 & 38.93 & 25.91 & 25.62 & 37.19 & 28.78 & 28.53 \\
        COD (PointPillar+RT-DETR)   & \ding{55}                         &66.59 & 58.72 & 53.6  & 52.59 & 43.93 & 37.83 & 42.02 & 27.57 & 26.33 & 53.73 & 43.41 & 39.25 \\  
        COD (PointPillar+RT-DETR)   & \ding{51}                         &81.20  & 81.73 & 75.82 & 58.65 & 50.45 & 44.22 & 55.50  & 37.16 & 35.43 & 65.12 & 56.45 & 51.82 \\      
    \bottomrule
    \end{tabular}
    }
    \label{tab:comparison_kitti_noise}
\end{table*}

\begin{table*}[!t]
\caption{Ablation experiments for various types of noise on the KITTI dataset.}
\centering
\resizebox{\linewidth}{!}{
  \renewcommand{\arraystretch}{1.1} %
    \begin{tabular}{lccllllllllllll}
    \toprule
    \multirow{2}{*}{Detector Modality} & \multicolumn{2}{c}{Noise}                        & \multicolumn{3}{c}{Car (Bbox AP@0.7)} & \multicolumn{3}{c}{Pedestrian (Bbox AP@0.5)} & \multicolumn{3}{c}{Cyclist (Bbox AP@0.5)} & \multicolumn{3}{c}{Bbox mAP} \\
    \cmidrule(lr){2-3} 
    \cmidrule(lr){4-6}
    \cmidrule(lr){7-9}
    \cmidrule(lr){10-12}
    \cmidrule(lr){13-15}
                                       & rot.               & trans.                  & easy   & mod.   & hard  & easy     & mod.     & hard     & easy    & mod.    & hard    & easy   & mod.   & hard  \\
    \midrule
    CenterPoint~\cite{centerpoint}                        & \multirow{2}{*}{\ding{51}}   & \multirow{2}{*}{\ding{51}}    & 36.53  & 28.78  & 29.97 & 47.63    & 42.00    & 41.08    & 41.04   & 28.16   & 27.84   & 41.74  & 32.98  & 32.96 \\
    COD (CenterPoint+RT-DETR)                &     &                            & 60.23  & 54.67  & 50.72 & 58.08    & 51.55    & 45.36    & 52.69   & 36.41   & 31.00   & 57.00  & 47.54  & 42.36 \\
    \midrule
    CenterPoint~\cite{centerpoint}                        & \multirow{2}{*}{\ding{55}}   & \multirow{2}{*}{\ding{51}}    & 86.78  & 77.89  & 78.04 & 62.03    & 54.89    & 53.73    & 67.02   & 49.70   & 42.50   & 71.94  & 60.83  & 58.09 \\
    COD (CenterPoint+RT-DETR)                &    &                         & 79.44  & 80.81  & 74.80 & 61.63    & 54.82    & 53.31    & 58.89   & 41.24   & 40.95   & 66.65  & 58.96  & 56.35 \\
    \midrule
    CenterPoint~\cite{centerpoint}                         & \multirow{2}{*}{\ding{51}}   & \multirow{2}{*}{\ding{55}}         & 37.40  & 29.32  & 30.49 & 45.45    & 40.78    & 40.28    & 39.36   & 26.32   & 26.56   & 40.74  & 32.14  & 32.44 \\
    COD (CenterPoint+RT-DETR)               &    &                            & 60.25  & 54.87  & 50.88 & 56.83    & 50.52    & 44.32    & 51.13   & 30.10   & 29.85   & 56.07  & 45.17  & 41.68 \\
    \bottomrule
    \end{tabular}
    }
\label{tab:different_noise_kitti}
\end{table*}

\subsection{Main Results}
\subsubsection{Overall Performances}
The consistency detection method proposed in this article is capable of simultaneously detecting the positions of objects in both point clouds and images through a single forward inference, assigning the same ID to the same object across both modalities. In traditional 3D detection algorithms, the detection of image bounding boxes (Bbox) is calculated using a calibration matrix between the point cloud and the image, rather than being directly inferred by the network, making the precision of image Bbox highly dependent on the accuracy of the calibration matrix. The accuracy comparisons for 3D and 2D detections between traditional methods and the proposed consistency detection approach are presented in Table~\ref{tab:consistency_kitti} and Table~\ref{tab:consistency_v2x}. 
From Table~\ref{tab:consistency_kitti}, it is evident that the method proposed in this article demonstrates performance comparable to traditional methods in terms of both 3D detection accuracy and 2D Bbox precision. 
However, as depicted in Table~\ref{tab:consistency_v2x}, in the context of DAIR-V2X data, the precision of traditional methods' Bbox, which relies heavily on the accuracy of the calibration matrix between point clouds and images, degrades significantly when the matrix is imprecise.
This trend is further confirmed by the subsequent Table~\ref{tab:different_noise_kitti}. 
In contrast, the Bbox precision inferred by the proposed method remains high, achieving double the accuracy of the traditional methods. This illustrates the robustness of the proposed method against calibration errors in real-world scenarios.

\subsubsection{Consistency Precision Analysis}
We have evaluated the accuracy of target correspondence using the consistency method between images and point clouds and compared it to traditional match-based methods for determining correspondences. These traditional methods first use separate 3D and 2D detectors to identify targets and then employ the calibration matrix between point clouds and images to match 3D detection boxes with 2D detection boxes.

The results of the consistency detection are displayed in Table~\ref{tab:comparison_traditional_improved}. Here, ``3D detection \& RT-DETR'' describes the approach where the 3D detector and RT-DETR function independently to identify objects, which is followed by a post-processing phase that aligns the 2D and 3D detection outcomes. Conversely, ``3D detection+RT-DETR'' represents the consistency detection method introduced in this paper. The table distinctly illustrates that the consistency precision of the proposed method surpasses traditional methods, demonstrating enhanced performance even under noisy conditions. This underscores the robustness of the proposed consistency detection approach, particularly its capability to sustain high accuracy across diverse modalities in sub-optimal conditions.

\subsection{Ablation Studies}
\subsubsection{Noise Resistance of Different Methods}
Noise ablation experiments were conducted to evaluate the performance of the method proposed in this paper for image bounding box detection when the calibration matrix is inaccurate. These experiments compared traditional methods, the consistency detection method proposed in the paper, and the consistency detection method with noise introduced during training. In Table~\ref{tab:comparison_kitti_noise}, ``-'' indicates that the training process is unaffected by calibration matrix noise, ``\ding{55}'' denotes training without calibration matrix noise, and ``\ding{51}'' signifies training with noise in the calibration matrix. 

The final results, tested under conditions where the calibration matrix contained noise as shown in Table~\ref{tab:comparison_kitti_noise}, demonstrate that compared to traditional computational methods, the proposed method exhibits superior resistance to interference. Additionally, introducing noise during the training phase further enhances this interference resistance, demonstrating the effectiveness of noise-augmented training in improving the robustness of the detection system under conditions of calibration matrix inaccuracies.

\subsubsection{Different Noise Effects}
Different types of noise were introduced to compare our method with traditional methods in terms of image detection accuracy, further exploring the robustness of our approach under conditions of inaccurate camera parameters. 
In Table~\ref{tab:different_noise_kitti}, `rot.' and `trans.' represent rotational and translational noise, respectively, each with a mean value of zero and variances of $0.01$ and $0.002$, expressed in degrees and meters, respectively.

In Table~\ref{tab:different_noise_kitti}, we observe that the post-processing method is less resilient to translational interference compared to rotational interference. Specifically, our method performs excellently in the presence of combined noise and individual rotational noise, while exhibiting slightly inferior performance compared to traditional methods in the presence of translational noise. This is because the translational variance set by us is minimal, resulting in only slight shifts in the bounding boxes calculated using traditional methods, thereby minimally impacting the mean Average Precision (mAP). In contrast, rotational noise can cause significant deformation to the calculated bounding boxes, thus having a larger impact on mAP. Our method maintains better performance under both types of noise, demonstrating stronger robustness.

\section{Conclusion}
This paper has introduced the task of consistency detection and proposes a method to address it. 
Moreover, to measure the accuracy of consistency, a specific metric, Consistency Precision (CP), is introduced. Extensive experiments were conducted to demonstrate the effectiveness and robustness of the proposed method, thereby establishing a benchmark for consistency detection. Consistency detectors are instrumental in locating the same target across different modalities, which is of significant importance for human-machine interaction and environmental perception. It is anticipated that such technology will find widespread applications in various interactive fields in the future.

\bibliographystyle{IEEEtran}
\bibliography{IEEEexample}

\end{document}